\documentclass[10pt,twocolumn,letterpaper]{article}

\usepackage{iccv}              
\usepackage{amsmath,amsfonts}

\usepackage{amssymb}
\usepackage{algorithmic}
\usepackage{algorithm}
\usepackage{array}
\usepackage{textcomp}
\usepackage{stfloats}
\usepackage{float}
\usepackage{url}
\usepackage{verbatim}
\usepackage{graphicx}
\usepackage{booktabs}
\usepackage{bm}
\usepackage{pifont}
\usepackage{enumitem}
\usepackage{soul}
\usepackage{makecell}
\usepackage{tabularx}
\usepackage{multirow}
\usepackage{siunitx}
\usepackage{balance}
\usepackage{lscape}
\usepackage[utf8]{inputenc}
\usepackage[T1]{fontenc}
\usepackage{pmboxdraw} 
\usepackage{tocloft}
\usepackage{listings}
\usepackage{rotating}
\usepackage{makecell}
\usepackage{caption}
\usepackage{multicol}
\usepackage{wrapfig}
\usepackage{colortbl}
\usepackage{arydshln}

\usepackage{xspace}
\usepackage{rotating}
\usepackage{framed}
\usepackage{bbm}
\usepackage{gensymb}
\usepackage[export]{adjustbox}
\definecolor{iccvblue}{rgb}{0.21,0.49,0.74}
\usepackage[pagebackref,breaklinks,colorlinks,allcolors=iccvblue]{hyperref}
\usepackage[capitalize]{cleveref}

\crefname{section}{Sec.}{Secs.}
\Crefname{section}{Section}{Sections}
\Crefname{table}{Table}{Tables}
\crefname{table}{Tab.}{Tabs.}

\def \ie {\emph{i.e.},}
\def \eg {\emph{e.g.},}

\def \wrt {\emph{w.r.t.}}

\newcolumntype{Y}{>{\centering\arraybackslash}X}

\newcommand{\tit}[1]{\smallbreak\noindent\textbf{#1.}}

\title{Quo Vadis Handwritten Text Generation for Handwritten Text Recognition?\vspace{-.3cm}}

\author{
Vittorio Pippi$^{1}$ \quad Konstantina Nikolaidou$^{2}$ \quad Silvia Cascianelli$^1$ \quad George Retsinas$^{3}$ \\ \quad Giorgos Sfikas$^{4}$ \quad Rita Cucchiara$^1$ \quad Marcus Liwicki$^2$\vspace{.2cm} \\ 
\begin{tabular}{ccc}
\makecell{$^1$University of Modena and Reggio Emilia} & \makecell{$^2$Luleå University of Technology}\\
\makecell{$^3$National Technical University of Athens} & 
\makecell{$^4$University of West Attica}
\vspace{-.3cm}
\end{tabular}
}

\begin{document}
\maketitle
\begin{abstract}
The digitization of historical manuscripts presents significant challenges for Handwritten Text Recognition (HTR) systems, particularly when dealing with small, author-specific collections that diverge from the training data distributions. Handwritten Text Generation (HTG) techniques, which generate synthetic data tailored to specific handwriting styles, offer a promising solution to address these challenges. 
However, the effectiveness of various HTG models in enhancing HTR performance, especially in low-resource transcription settings, has not been thoroughly evaluated. 
In this work, we systematically compare three state-of-the-art styled HTG models (representing the generative adversarial, diffusion, and autoregressive paradigms for HTG) to assess their impact on HTR fine-tuning. 
We analyze how visual and linguistic characteristics of synthetic data influence fine-tuning outcomes and provide quantitative guidelines for selecting the most effective HTG model. The results of our analysis provide insights into the current capabilities of HTG methods and highlight key areas for further improvement in their application to low-resource HTR.
\end{abstract}    
\section{Introduction}
\label{sec:intro}
The process of digitizing documents is becoming essential across both cultural and industrial sectors for their effective management~\cite{quattrini2024mu}, preservation~\cite{cascianelli2022lam}, and enhancement~\cite{quattrini2023volumetric,quattrini2024binarizing}. As a result, Document Analysis tasks, particularly those focused on handwritten text~\cite{pippi2023evaluating}, have been attracting great attention from the research community. 
Modern Handwritten Text Recognition (HTR) systems~\cite{kang2022pay,retsinas2022best,retsinas2021seq2seq,retsinas2024enhancing}, which are typically trained on large publicly available datasets, perform well on documents that resemble the data used for training. However, their performance significantly declines when tested on documents that differ substantially from the training data. 
A key challenge arises with historical manuscripts~\cite{nikolaidou2022survey} held in archives. 
These are usually small but valuable collections that often feature limited pages written by specific, historically important authors. These manuscripts display unique stylistic and linguistic features that pose difficulties for current HTR systems. 
To address this challenge, developing strategies that optimize HTR performance for such materials is critical for their efficient digitization.
A common approach involves pretraining HTR models on large-scale datasets, either real or synthetic, followed by fine-tuning them on a small set of real data from the target domain~\cite{pippi2023choose}. 
Some research work has already been devoted to exploring the use of synthetic datasets for pretraining HTR systems~\cite{granet2018transfer, aradillas2020boosting, kang2022pay, wick2021rescoring, li2021trocr}. The effectiveness of these strategies largely depends on the extent to which the synthetic data mirrors real-world data~\cite{cascianelli2021learning,pippi2023choose}. In response, Handwritten Text Generation (HTG) techniques, particularly styled HTG, have emerged as promising tools~\cite{bhunia2021handwriting,pippi2023handwritten,vanherle2024vatr++, nikolaidou2024diffusionpen,pippi2025zero}. These models allow for the generation of training data tailored to specific domains by synthesizing images of text in a desired handwriting style by using just a few sample images as a reference. Styled HTG models typically include an encoder to extract the style features from the examples and a generator that combines these features with a desired text representation to produce text images with control over style and content.

Recent years have witnessed the development of various HTG paradigms with great improvements in terms of reference style fidelity, making them potentially very useful for generating tailored training data for HTR models~\cite{nikolaidou2024rethinking}. 
Nonetheless, a systematic evaluation of such usefulness in low-resource HTR scenarios is still missing in the literature. 
In light of this, in this paper, we explore a pretraining with additional fine-tuning strategy based on automatically generated, author-specific synthetic datasets for comparing styled HTG networks, each one being the state-of-the-art and representing the three main existing paradigms for tackling the taks: a generative adversarial model~\cite{vanherle2024vatr++}, a diffusion model~\cite{nikolaidou2024diffusionpen}, and an autoregressive model~\cite{pippi2025zero}. 
Via extensive evaluation, we assess the effectiveness of these HTG approaches when generating pretraining data for HTR scenarios spanning multiple languages, various authors, and different historical periods. 
The results of our analysis give insights into the current capabilities of HTG models and suggest key areas for future research to improve their applicability in low-resource HTR scenarios.
\section{Related Work}
\label{sec:related}

HTR is a well-established area of research due to its wide range of applications in both industrial and cultural sectors. Despite its progress, HTR remains a complex and challenging problem. The task can be tackled at different levels of granularity, ranging from individual characters, often used for idiomatic languages~\cite{cilia2019ranking}, to entire words~\cite{such2018fully, bhunia2019handwriting}, lines~\cite{shi2016end, puigcerver2017multidimensional}, paragraphs, and full pages~\cite{moysset2017full, bluche2017scan, bluche2016joint, wigington2018start, clanuwat2019kuronet}. Line-level recognition is particularly common for non-idiomatic languages, where it can be applied as a standalone method or integrated into a broader page-level system~\cite{yousef2020origaminet, moysset2017full, bluche2017scan}. The majority of modern HTR systems employ learning-based approaches, relying on Multi-Dimensional Long Short-Term Memory networks (MD-LSTMs)~\cite{shi2016end, puigcerver2017multidimensional, pham2014dropout, voigtlaender2016handwriting, bluche2017gated} for feature extraction. 
These methods typically use the Connectionist Temporal Classification (CTC) decoding strategy to produce text transcriptions~\cite{graves2009offline, bluche2016joint}. 
Recently, alternative models based on fully convolutional networks~\cite{yousef2020origaminet, coquenet2020recurrence} and Transformer encoder-decoder architectures~\cite{kang2022pay, li2021trocr, wick2021transformer} have also been proposed for HTR tasks~\cite{vaswani2017attention}.
To improve transcription quality, explicit language models or lexicons can be employed. However, the effectiveness of these models depends on the consistency and regularity of the transcribed language, especially regarding the presence of rare words, proper nouns, or errors. This makes language models less reliable, particularly when working with historical manuscripts where the language can be highly variable or archaic. 
In this work, we focus on line-level HTR and historical data. 
Thus, we do not rely on any lexicon or explicit language model.

A significant challenge in HTR is the scarcity of training data, particularly for single-author documents or ancient manuscripts with unique characteristics. One solution to address this limitation is data augmentation, which can involve general image manipulations such as color changes and geometric transformations~\cite{voigtlaender2016handwriting, puigcerver2017multidimensional, wigington2017data} or more targeted modifications specifically designed to reflect the characteristics of the target data~\cite{chammas2018handwriting}.
Another widely adopted method is pretraining the HTR model on large, diverse datasets followed by fine-tuning on a smaller set of target-specific data~\cite{granet2018transfer, jaramillo2018boosting, soullard2019improving}. This approach has been demonstrated to outperform basic data augmentation when applied to historical manuscripts~\cite{aradillas2020boosting}. The pretraining data can consist of real handwritten text \eg~publicly available benchmark datasets, or synthetic data, often generated by rendering text in calligraphic fonts~\cite{shen2016method, kang2022pay,cascianelli2021learning}. 
For single-author scenarios,~\cite{pippi2023choose} highlights the importance of considering both the overall appearance (such as the type of paper, writing instrument, and average character width) and the language (including the time period and the topic) when selecting real datasets or generating synthetic ones for pretraining. Additionally, they demonstrate that an HTR model trained on text images with a wide range of handwriting styles tends to be more adaptable and robust compared to one trained on a single handwriting style. However, when synthetic data closely mimic the actual handwriting in the real data, achieving satisfactory performance becomes feasible.

Recent research has explored using HTG models to generate synthetic data for training HTR models, aiming to enhance their performance on real-world datasets~\cite{souibgui2022one,kang2021content,nikolaidou2023wordstylist,pippi2023choose}. 
HTG entails generating realistic handwritten text images. 
In its styled variant, which is the focus of this work, the goal is to produce writer-specific handwritten text by using just a few example images to capture and replicate the writer's style~\cite{bhunia2021handwriting,fogel2020scrabblegan,kang2020ganwriting}.
Three main paradigms have been proposed to tackle this task. 
The predominant technique is the use of generative-adversarial models~\cite{kang2020ganwriting, bhunia2021handwriting, pippi2023handwritten, vanherle2024vatr++,gan2021higan, gan2022higan+, krishnan2023textstylebrush}. 
A few works have explored HTG using Diffusion Models~\cite{luhman2020diffusion,zhu2023conditional,nikolaidou2023wordstylist,ding2023improving}, which led to impressive performance. 
Finally, a recent work introduced an autoregressive approach to HTG~\cite{pippi2025zero} using only synthetic data for training. 
In this work, we consider a representative HTG model for each of these paradigms and evaluate them in the context of low-resource HTR.
\section{Method}
\label{sec:approach}

\begin{figure*}[t]
    \centering
    \includegraphics[width=\textwidth]{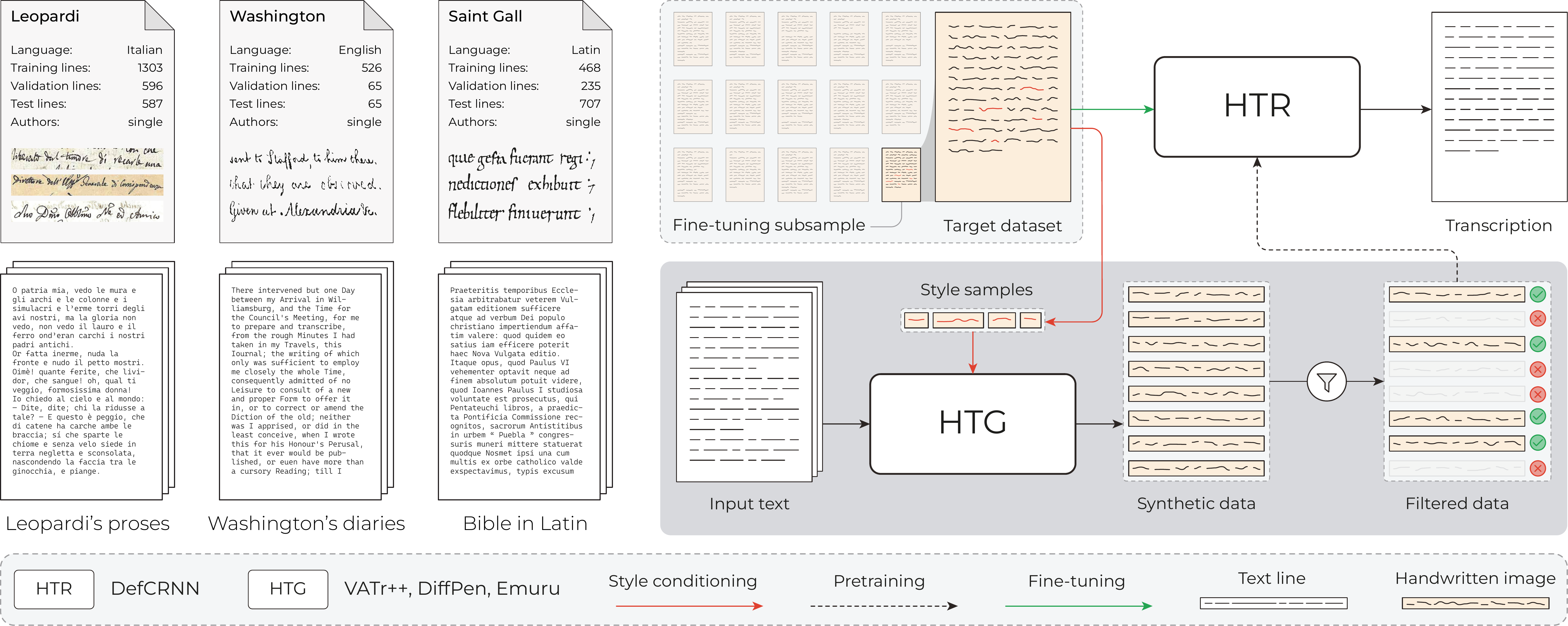}
    \caption{Overview of our pipeline for synthetic data generation from collection-specific handwritten lines. The generation process renders line images from a given text conditioned by a few style samples from the target dataset. Then, the synthetic dataset is filtered based on readability or fidelity criteria and is used to pre-train the recognition model before possible fine-tuning on the real data.}
    \label{fig:overview}
\end{figure*}

In this work, we aim to evaluate the performance of state-of-the-art off-the-shelf HTG models when used to generate synthetic pretraining datasets for HTR on small collections of documents with distinctive characteristics (\eg~unique handwriting styles, language variations, paper supports).
Moreover, we investigate strategies to maximize the benefit of using such models. 
To these ends, we devise a pipeline that entails synthetic data generation and subsequent HTR pretraining, followed by a final fine-tuning on limited real data.
In particular, we pretrain an HTR model on a large synthetic dataset obtained from HTG models to imitate the style of a real target dataset, followed by filtering strategies based either on the readability or style fidelity of the generated samples. 
Finally, the pipeline entails fine-tuning using a limited number of real samples from the target collection. 
To create the synthetic datasets, we require exemplar style images, which can be easily extracted from digitized manuscripts within the target collection. 
Additionally, the textual content to be rendered in the desired handwriting style must be specified.
We consider the scenario where the author and the language of the manuscript are known. 
In this case, if transcribed texts written by the author of interest exist, we can use the HTG model to generate synthetic samples of these texts. 
Otherwise, if only the manuscript’s language is known (or if no existing texts by the same author are available), the HTG model can generate text in the same language as the target collection. 
In both cases, the HTG model outputs handwritten line images of varying lengths. Note that the quality of some of the generated lines can be non-ideal, limiting their usefulness for HTR training. 
To limit this risk, a possible solution is to filter out synthetic images that do not meet certain quality criteria. 
In this work, we consider two alternatives: the line readability, expressed in terms of Character Error Rate (CER) of an off-the-shelf, language-agnostic HTR model, and the style fidelity \wrt~the reference, expressed in terms of Handwriting Distance (HWD)~\cite{pippi2023hwd}.

In the following, we present the main components of our proposed pipeline (see~\Cref{fig:overview}).
We describe the HTR model used for transcription (dubbed \textbf{DefCRNN}~\cite{cascianelli2022boosting}) and the HTG models considered to generate synthetic pretraining data. These are the generative-adversarial Transformer \textbf{VATr++}~\cite{vanherle2024vatr++}, the diffusion-based \textbf{DiffusionPen (DiffPen)}~\cite{nikolaidou2024diffusionpen}, and the autoregressive generative Transformer \textbf{Emuru}~\cite{pippi2025zero}. Finally, we give the details of the proposed \textbf{HWD}-based and \textbf{CER}-based filtering strategies. 

\subsection{HTR Approach}
The combination of convolutional and recurrent neural networks has long been a standard approach for HTR, and it is widely used due to its effectiveness and efficiency. 
In this work, we employ a model based on one-dimensional LSTM networks, which offer comparable or even superior performance compared to MD-LSTMs~\cite{puigcerver2017multidimensional}.
Our model is based on a variant of the CRNN method~\cite{shi2016end}, referred to as DefCRNN~\cite{cascianelli2022boosting}. The convolutional part of the architecture consists of seven convolutional blocks. The first six blocks follow the VGG-11 structure, with modifications to the final two max-pooling layers to incorporate rectangular pooling, which helps preserve the aspect ratio of text line images. The seventh convolutional block utilizes a $2\times2$ kernel. The variant we exploit contains Deformable Convolutions~\cite{dai2017deformable} as proposed in~\cite{cojocaru2020watch, cascianelli2021learning, cascianelli2022boosting}, which enhance model performance by allowing for more flexible feature extraction.
The output of the final convolutional layer is a feature map of size $2 \times W \times 512$, where $W$ is determined by the width of the input image. This feature map is then collapsed along the channel dimension, resulting in a sequence of $W$ feature vectors, each with a size of 1024. 
These vectors are passed to the recurrent module, which consists of two Bidirectional LSTM layers with 512 hidden units each, with a dropout layer (probability 0.5) in between. 
The output of the recurrent module is a sequence of probability distributions over character classes for each feature vector.
As is typical in HTR, the model is trained using the CTC loss function, which includes a special \textit{blank} character. 
Notably, we do not use any external language model to ensure that the model is adaptable across different languages.

\subsection{HTG Approaches}
Styled HTG models efficiently create large volumes of synthetic text images in a specified handwriting style, starting from a few real images from the target dataset. 
An overview of the considered HTG approaches, each representative of a distinct paradigm among adversarial-, diffusion-, and autoregressive-based, is reported below (we refer the interested reader to the respective papers for more details). 
To explore the full potential of the HTG models, we use them in an off-the-shelf manner, as proposed in the original works,  without training on the target datasets considered.

\tit{VATr++}
VATr++~\cite{vanherle2024vatr++} employs a generator-discriminator framework~\cite{goodfellow2014generative, mirza2014conditional}, complemented by an auxiliary HTR network for readability and a writer classification module to ensure stylistic fidelity.
The model has been designed to address the generation of rare or out-of-charset characters. 
This is achieved by encoding target text as a sequence of Visual Archetypes (VAs)~\cite{pippi2023handwritten}, which allow the model to exploit geometric similarities between glyphs and by adopting specific training and data preparation strategies.
The architecture is a hybrid Convolutional-Transformer encoder-decoder. 
The encoder uses a synthetically pre-trained CNN to process the reference style samples, while the Transformer encoder aggregates them into a style vector using self-attention. 
The Transformer decoder aligns this representation with a sequence of VAs representing the desired text content through cross-attention, and a convolutional decoder synthesizes the final handwritten image. 
VATr++ accepts as input 15 word images, from which it extracts the style and generates word or text line images.

\tit{DiffPen} 
DiffPen~\cite{nikolaidou2024diffusionpen} is a latent diffusion model that synthesizes images conditioned on a text prompt and style features in a few-shot setting~\cite{quattrini2024merging,quattrini2024alfie}. 
Similar to standard conditional latent diffusion models~\cite{rombach2022high}, the method utilizes a U-Net-based architecture~\cite{ronneberger2015u} as the backbone denoising network, and a pre-trained Variational Autoencoder (VAE)~\cite{kingma2014auto} to encode and decode images from pixel to latent space and vice-versa.
Two auxiliary pre-trained encoders are used for the text and style conditions.
To create the content embedding, an off-the-shelf pre-trained text encoder~\cite{clark2022canine} operates on character sequences.
As the style encoder, the work proposes a CNN feature extractor that combines classification and metric learning to construct a continuous embedding space that supports diverse output and enables fine-grained control (\eg~style interpolation and mixing). 
To create the style condition, DiffPen extracts features from 5 style examples of the same writer and generates word images with the desired content. 
Although DiffPen is designed for word-level generation, the authors have proposed patching together subparts of text or words to obtain longer text or complete lines, which we also adopt in our work.

\tit{Emuru} 
Emuru~\cite{pippi2025zero} is a continuous-token autoregressive model for handwritten text generation, capable of producing text images of any length while preserving style fidelity and readability. 
It enhances generalization to novel styles and minimizes background artifacts. 
The architecture consists of a VAE~\cite{kingma2014auto} and an autoregressive Transformer Encoder-Decoder. 
The VAE maps style reference images into a continuous latent space, encoding only the writing style while removing background noise. 
The Transformer takes as input the style embeddings, the text present in the reference image, and the desired text, iteratively generating an image that preserves the target style.
Both components of Emuru are trained on a large synthetic dataset. 
This ensures that the VAE learns to reconstruct text without background artifacts while providing a style representation that generalizes well to new handwriting styles and typefaces. 
The model generates text images in an autoregressive loop, where each iteration outputs visual embeddings that are then decoded by the VAE into a final styled image. 
This iterative process allows the model to determine its own stopping point, eliminating constraints on maximum text length.
Emuru takes as input a text line image with its associated text content and is designed to generate entire text lines.

\subsection{Filtering Approaches}
We argue that not all the generated samples are equally useful and of high quality for HTR pretraining. 
Despite their impressive performance, HTG models tend to generate either noisy data that might not preserve the content or simplify the generated style, reducing variability~\cite{nikolaidou2024rethinking}.
This could negatively impact the performance of an HTR system when faulty content or data with low variability are integrated into the training process.
To explore this, we propose an analysis based on two different criteria (\ie~readability and style fidelity) and discard those that do not meet a predefined quality threshold. 
In the following parts, we describe the considered filtering strategies. 

\tit{Readability} 
To evaluate how the readability of the generated samples affects the training of the HTR model, we measure the CER for each synthetic image by using a pretrained TrOCR network~\cite{li2021trocr} and then filter out those for which the CER is above a certain threshold. 
We define four filtering thresholds (\ie~CER$<$0.15, CER$<$0.30, CER$<$0.45, and CER$<$0.60), which progressively include samples based on transcription quality. 
Images that satisfy stricter thresholds are considered more readable, as they exhibit fewer transcription errors. Conversely, images that exceed higher CER values are discarded during the filtering process and will not be used to pre-train the DefCRNN.

\tit{Style Fidelity} 
To assess the stylistic fidelity of the generated samples, we quantify how closely each synthetic image matches the handwriting style of the real ones. 
We compute the HWD~\cite{pippi2023hwd} between each generated image and a representative style embedding extracted from the real samples. 
This embedding is obtained by averaging the features of real handwriting samples, serving as a reference for style similarity.
Since each of the HTG models produces a distinct distribution of HWD values, we define filtering thresholds using the 25th, 50th, and 75th percentiles of its respective HWD distribution. 
Samples with HWD below a given percentile are considered more stylistically faithful, while those above the higher percentiles are progressively filtered out. 
This percentile-based approach ensures that style fidelity is evaluated fairly according to each model’s intrinsic variability in the generated handwriting.

\section{Experiments}
\label{sec:experiments}

\subsection{HTR Training Details}
\label{sec:implementation_details}
For training the DefCRNN model, all input images are rescaled to a height of 64 pixels while maintaining their original aspect ratio, followed by intensity normalization to the range $[-1,1]$. 
During pretraining, we apply a series of augmentations to enhance robustness. Brightness is adjusted using a randomly sampled factor from $[0.5, 5]$, contrast from $[0.1, 10]$, saturation from $[0, 5]$, and hue from $[-0.1, 0.1]$. 
Additionally, Gaussian blur with a kernel size of 5 is applied, with a standard deviation randomly chosen from $[0.1, 2]$. To introduce geometric variability, we randomly apply one of the following transformations: a slight rotation between $-1\degree$ and $1\degree$, an affine transformation with rotation in the same range and shear between $-50\degree$ and $30\degree$, or a random tomography. 
Pretraining is conducted with a batch size of 16, which is reduced to 8 for fine-tuning and training from scratch. The model is optimized using Adam with $\beta_1 = 0.9$ and $\beta_2 = 0.999$, and a learning rate of $10^{-4}$ across all experiments. A scheduler reduces the learning rate by $10\%$ if the CER on the validation set plateaus. Early stopping is applied with a patience of 20 epochs, using CER as the criterion. 
When fine-tuning, optimal CER values are typically reached within the first few epochs, which usually require less than one hour.

\subsection{Datasets}
Our analysis considers three low-resources, line-level datasets as target collections. All of them are obtained from historical, single-author manuscripts. When generating synthetic data for pretraining, we consider the characteristics of each target dataset separately. 

\tit{Leopardi}
The Leopardi dataset~\cite{cascianelli2021learning} comprises a collection of early 19\textsuperscript{th}-century Italian manuscripts authored by Giacomo Leopardi, a prominent Romantic-era philologist, writer, and poet. It consists of 1303 training lines, 596 validation lines, and 587 test lines. All samples are RGB scans of ink-written texts on historical paper.

\tit{Washington}
The George Washington dataset~\cite{fischer2012lexicon} includes 20 handwritten English letters from 1755, authored by George Washington, the first U.S. President, and a collaborator. 
It is structured into 526 training lines, 65 validation lines, and 65 test lines. The dataset consists of binarized images of these historical documents.

\tit{Saint Gall}
The Saint Gall dataset~\cite{fischer2011transcription} originates from a late 9\textsuperscript{th}-century Latin manuscript written by a single scribe. It spans 60 pages and is divided into 468 training lines, 235 validation lines, and 707 test lines. All images are binarized scans of the original manuscript pages.

\tit{Synthetic Data}
Following~\cite{pippi2023choose}, we employ the considered HTG models to generate synthetic samples tailored to each target dataset. 
This process involves conditioning the generation on a subset of the original dataset’s samples. Specifically, VATr++ and DiffPen use crops from 15 and 5 randomly selected line images, respectively, while Emuru operates with a single line reference. The textual content for the synthetic data is chosen to align with each dataset: excerpts from Giacomo Leopardi’s prose for the Leopardi dataset, passages from George Washington’s diaries for Washington, and a medieval Latin Bible for Saint Gall. This selection ensures linguistic consistency between the synthetic and real data.

\subsection{Evaluation Protocol}
To analyze the impact of pretraining and fine-tuning in scenarios where only a small portion of the target dataset is annotated, we fine-tune models on progressively smaller subsets of the training data. Specifically, we consider fractions of $100\%$, $50\%$, $25\%$, $10\%$, $5\%$, $2.5\%$, and $1.25\%$ of the available training lines. For comparison, we also train models from scratch using the same subsets. Moreover, we assess direct transfer by applying pretrained models to the target datasets without fine-tuning. 
 
To compare the considered HTG models, we first consider their generation performance, expressed in terms of multiple commonly applied scores. Specifically, these are: Fréchet Inception Distance (\textbf{FID})~\cite{heusel2017gans}, Kernel Inception Distance (\textbf{KID})~\cite{binkowski2018demystifying}, \textbf{HWD}~\cite{pippi2023hwd}, the binarized version of FID and KID (dubbed \textbf{BFID} and \textbf{BKID}, respectively), obtained by computing the scores on binarized images, and the Absolute Difference in the CER (\textbf{$\Delta$CER}) of the off-the-shelf TrOCR-Base~\cite{li2021trocr} model on the reference and generated images. 
Moreover, since the main goal of this work is to compare HTG approaches in terms of their effectiveness in providing synthetic data for HTR, we consider the recognition performance of the considered DefCRNN trained on such data. We report the performance in terms of CER, which is standard for text recognition.

\begin{figure*}[t]
    \centering
    \includegraphics[width=\textwidth]{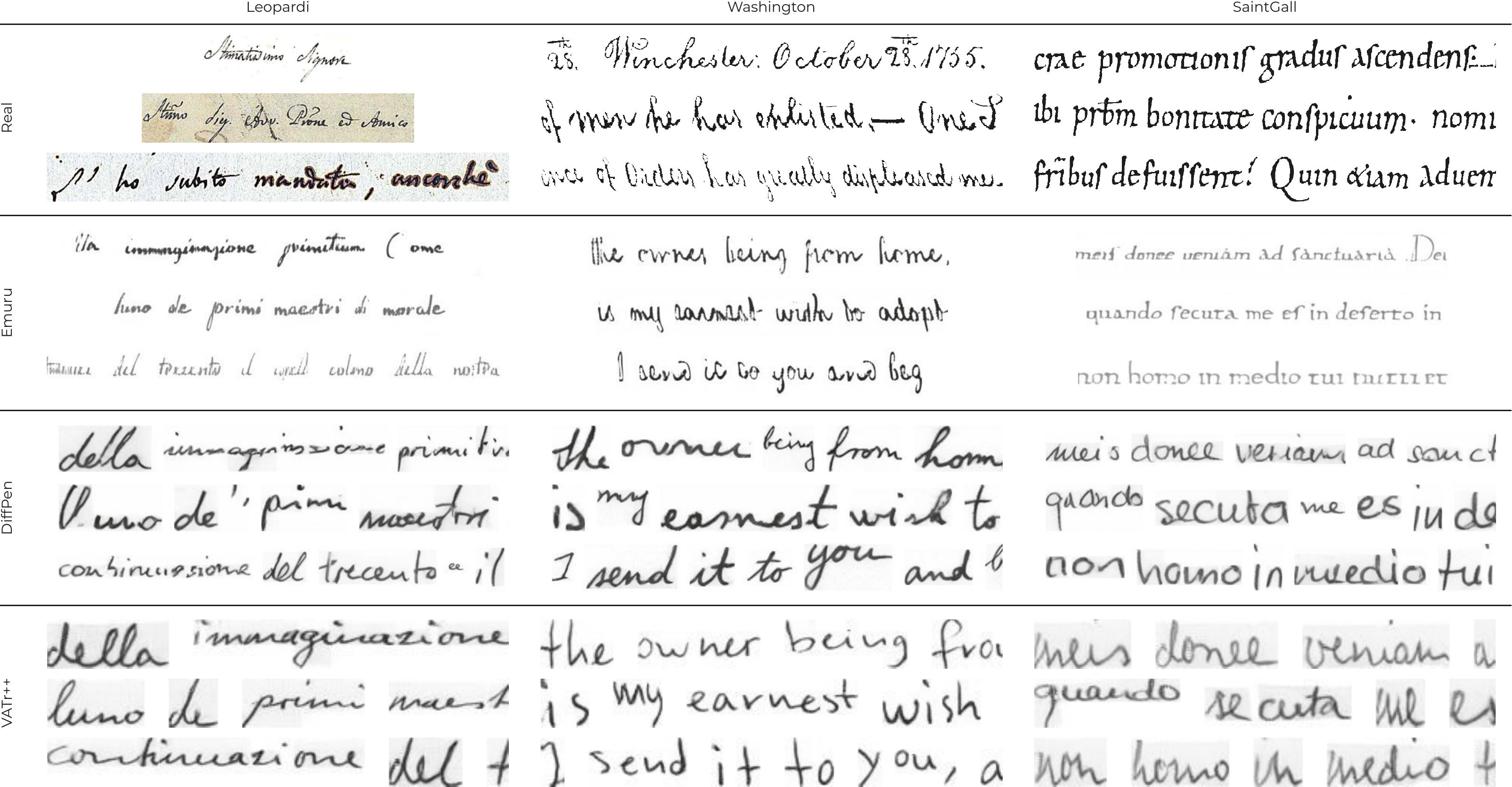}
    \caption{Qualitative comparison of the considered HTG models when generating samples from the considered target datasets. Note that none of the considered models has been trained on the target datasets.}
    \label{fig:datasets_samples}
\end{figure*}

\subsection{Results}

\begin{figure*}[t]
    \centering
    \begin{tabular}{ccc}
    \includegraphics[width=0.3125\textwidth]{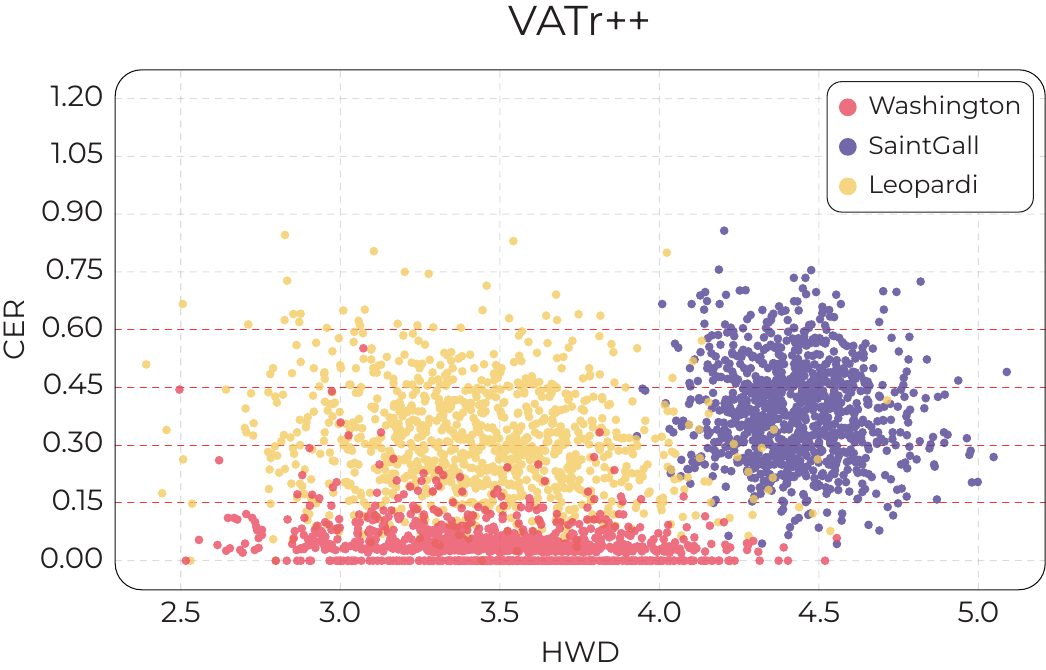}&
    \includegraphics[width=0.3125\textwidth]{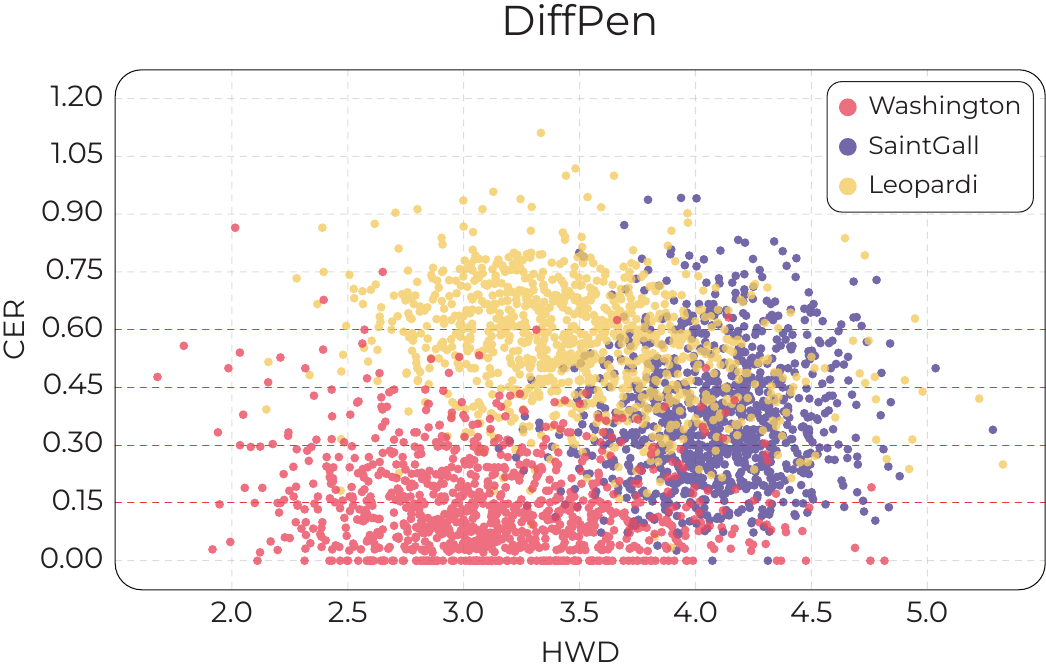}&
    \includegraphics[width=0.3125\textwidth]{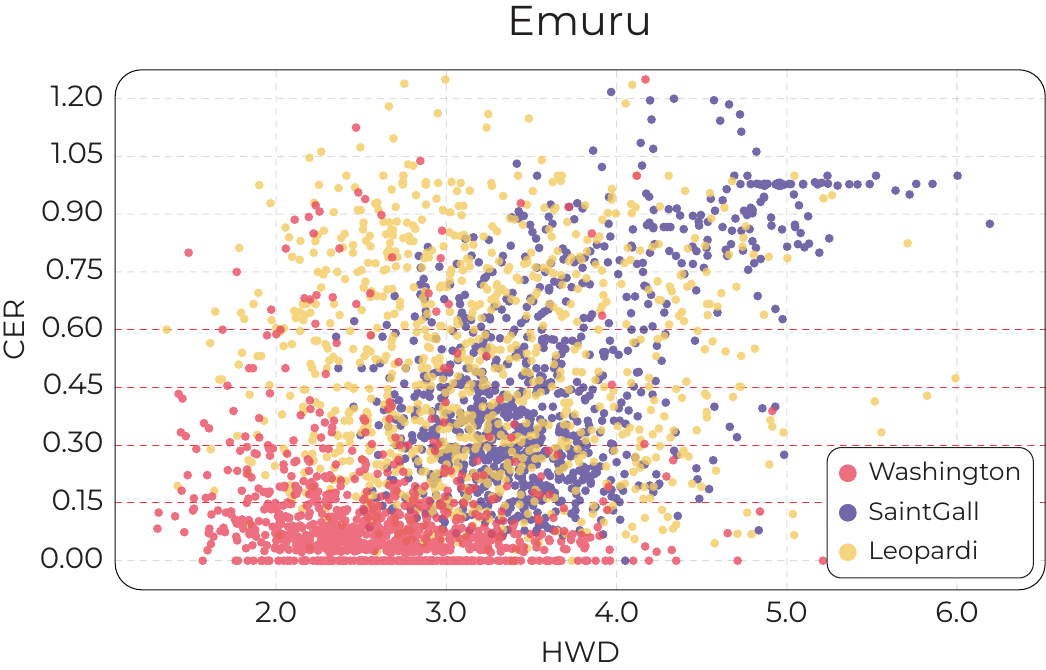}
    \end{tabular}
    \caption{Distribution of 1000 random samples from each synthetic dataset generated by the HTG models, in terms of CER and HWD \wrt~the real samples. The horizontal lines indicate the CER thresholds used for filtering. For readability, we omit the separators for HWD-based filtering since they depend on the percentiles.}
    \label{fig:scatters}
\end{figure*}

\tit{Generation Performance} 
First, we evaluate the considered HTG models in terms of their generation capabilities. Recall that none of them have been trained on the target datasets, making this a zero-shot evaluation of their ability to produce handwriting samples that align with the target styles. The quantitative performance comparison is reported in~\Cref{tab:generation}.
From the FID, KID, and HWD scores, it is evident that Emuru consistently outperforms the other models across nearly all metrics, leveraging its zero-shot capabilities to generate more style-faithful samples. This observation is confirmed by the qualitative examples in~\Cref{fig:datasets_samples}. 
Additionally, the scatter plots in~\Cref{fig:scatters}, which report the distribution of the generated datasets in terms of TrOCR CER and HWD relative to the respected target dataset, show that the images generated by Emuru exhibit the lowest readability according to the TrOCR model. However, from the $\Delta$CER values in~\Cref{tab:generation}, we can argue that this reduced readability is a consequence of Emuru’s ability to faithfully replicate the target handwriting style.
\begin{table}[]
\centering
\setlength{\tabcolsep}{.7em}
\resizebox{\linewidth}{!}{%
\begin{tabular}{ll cccccc}
\toprule
\textbf{Dataset} & \textbf{Model} & \textbf{HWD$\downarrow$} & \textbf{FID$\downarrow$} & \textbf{BFID$\downarrow$} & \textbf{KID$\downarrow$} & \textbf{BKID$\downarrow$} & \textbf{$\Delta\text{CER}\downarrow$} \\
\midrule
\multirow{3}{5em}{{Leopardi}}
& Emuru   & \textbf{2.05} & \textbf{208.4} & \textbf{~57.7} & 0.246 & \textbf{0.051} & \textbf{~0.4} \\
& DiffPen & 3.15 & 244.9 & 127.4 & 0.257 & 0.110 & ~1.6 \\
& VATr++  & 3.02 & 217.7 & ~90.8 & \textbf{0.230} & 0.084 & 21.8 \\
\midrule
\multirow{3}{5em}{{Washington}}
& Emuru   & \textbf{1.63} & \textbf{108.3} & \textbf{~24.6} & \textbf{0.104} & \textbf{0.016} & \textbf{~9.7} \\
& DiffPen & 2.56 & 171.6 & ~91.9 & 0.163 & 0.085 & ~9.8 \\
& VATr++  & 3.24 & 217.1 & 105.4 & 0.228 & 0.092 & 20.4 \\
\midrule
\multirow{3}{5em}{{Saint Gall}}
& Emuru   & \textbf{1.53} & \textbf{243.0} & \textbf{~32.2} & 0.320 & \textbf{0.014} & \textbf{~7.0} \\
& DiffPen & 3.44 & 250.0 & ~40.9 & 0.310 & 0.024 & 11.1 \\
& VATr++  & 3.82 & 251.6 & ~73.0 & \textbf{0.306} & 0.060 & 10.2 \\
\bottomrule
\end{tabular}%
}
\caption{Generation scores computed on the three target datasets generated with the considered HTG models.}
\label{tab:generation}\vspace{-0.5cm}
\end{table}

\begin{figure*}[t]
\centering
    \begin{tabular}{ccc}
    \includegraphics[width=0.3125\textwidth]{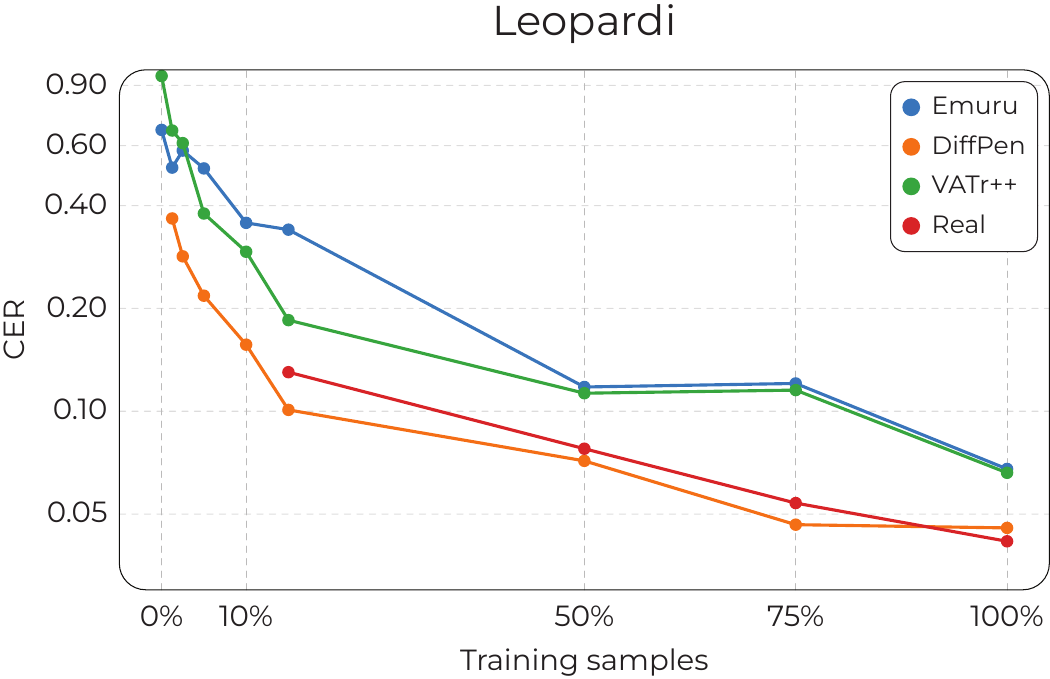}&
    \includegraphics[width=0.3125\textwidth]{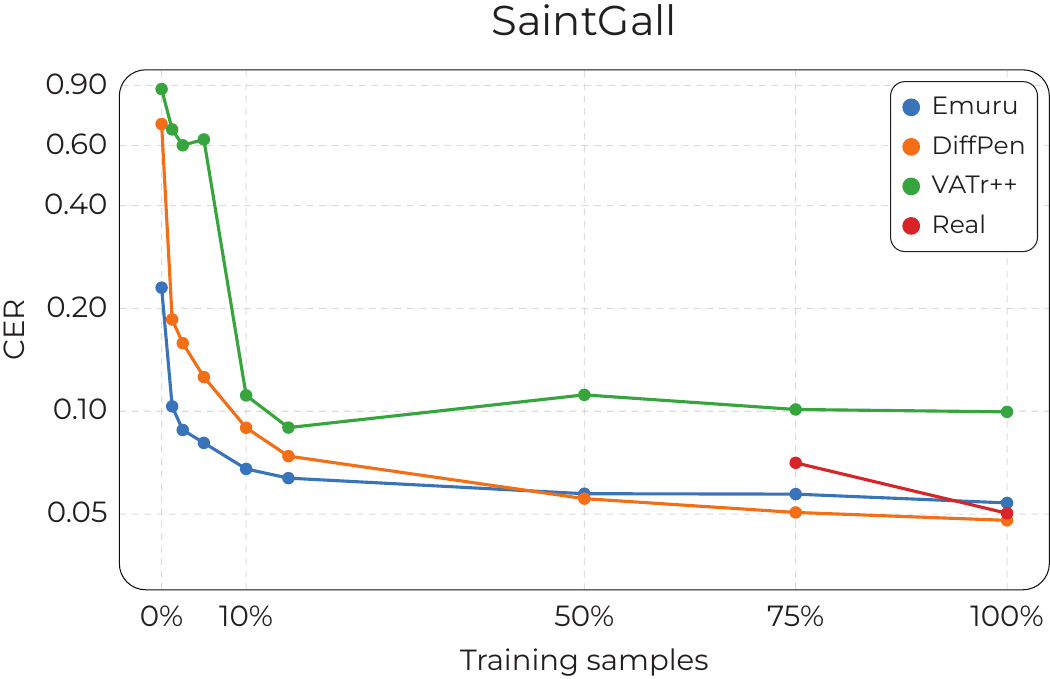}&
    \includegraphics[width=0.3125\textwidth]{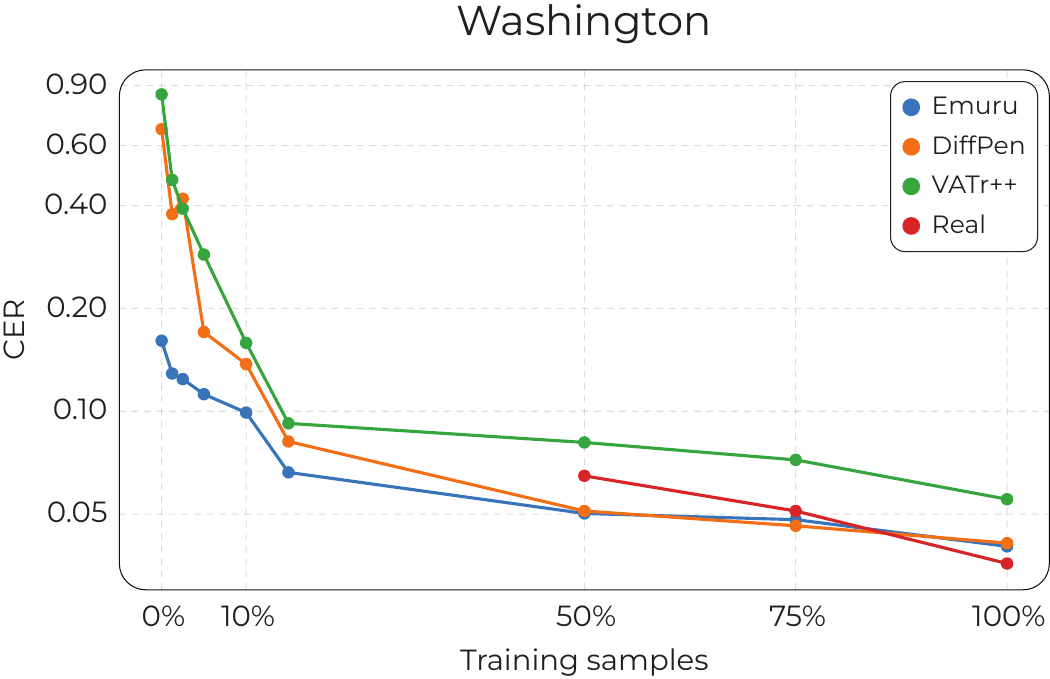}
    \end{tabular}
    \caption{CER scores obtained by fine-tuning DefCRNN on portions of the three target datasets, after having been pretrained on all the synthetic data generated with the considered HTG models. We report the CER obtained when training only on real data for comparison.}
\label{fig:scores_plot}\vspace{-0.2cm}
\end{figure*}

\tit{Direct Transfer Recognition Performance}
The recognition performance achievable by the DefCRNN model when pretrained on the generated data and then directly applied to the real target datasets are reported in~\Cref{tab:cer_leopardi,tab:cer_saintgall,tab:cer_washington} (in the first column relative to the CER) and depicted in~\Cref{fig:scores_plot}. As can be observed, Emuru's generated samples allow for achieving the best performance in this zero-shot HTR scenario, consistently outperforming the other models. 
A possible explanation for this advantage can be found in the scatter plots in~\Cref{fig:scatters}, where the generated samples from Emuru exhibit greater variability, forming more dispersed clusters. This diversity may contribute to the model's robustness when directly applied to unseen handwriting styles.
Furthermore, it is worth noting that the handwriting in the Washington and Saint Gall datasets is quite regular (see~\Cref{fig:datasets_samples}). Since Emuru is trained on a large corpus of synthetic typewritten and calligraphic fonts, it can effectively approximate the structured styles characteristic of these datasets. This alignment is reflected in the CER scores achieved in the direct transfer setting: 18.1 for Saint Gall and 15.3 for Washington.

\tit{Fine-tuning Recognition Performance}
From the results in~\Cref{tab:cer_leopardi,tab:cer_saintgall,tab:cer_washington,fig:scores_plot}, it can also be observed the effect of fine-tuning DefCRNN on a varying number of real data from the target dataset after being pretrained on the samples synthesized by the considered HTG models. It can be observed that Emuru is the most effective in providing pretraining data for the HTR model when only a very limited amount of real data is available for fine-tuning. In particular, when fewer than 130 real images are used (\eg~10\% of Leopardi, which corresponds to 130 images, or 25\% of Saint Gall, which includes 125 images), the style similarity between the pretraining and target dataset plays a crucial role. In other words, the closer the synthetic samples are to the target handwriting style, the greater the benefit for HTR performance in low-data regimes. 
However, as the number of real training samples increases beyond this threshold, the influence of style similarity of the pretraining dataset diminishes, and its diversity becomes the dominant factor in improving HTR performance. This explains why, when more than 130 real images are available, DefCRNN pretrained on the more stylistically varied DiffPen-generated datasets has better performance than when pretrained on Emuru-generated data. 
In other words, when more fine-tuning data are available, DiffPen's style variability provides a stronger generalization capability to the HTR model.

\begin{table}[]
\centering
\setlength{\tabcolsep}{.3em}
\resizebox{\linewidth}{!}{
\begin{tabular}{llc ccccccccc}
\toprule
    & & & \multicolumn{9}{c}{\textbf{CER ($\times$100)} } \\
\cmidrule{4-12}
\textbf{Data} & \textbf{Filter} & \textbf{\#samples} & \textbf{0\%} & \textbf{1.25\%} & \textbf{2.5\%} & \textbf{5\%} & \textbf{10\%} & \textbf{25\%} & \textbf{50\%} & \textbf{75\%} & \textbf{100\%} \\
\midrule
Real & -                    & ~1.3K & ~~~- & ~~~- & ~~~- & ~~~- & ~~~- & 13.0 & ~7.8 & ~5.4 & ~4.2 \\
\midrule
Emuru & $\text{HWD}_{25\%}$ & 21.5K & 62.7 & 48.6 & 47.4 & 26.9 & 20.4 & 12.9 & ~8.4 & ~8.0 & ~4.1 \\
Emuru & $\text{HWD}_{50\%}$ & 43.0K & \underline{\textbf{49.0}} & 34.9 & 29.7 & 23.8 & 15.2 & \underline{\textbf{10.0}} & \underline{~7.7} & \underline{~5.5} & ~4.6 \\
Emuru & $\text{HWD}_{75\%}$ & 64.5K & 49.3 & 45.5 & 52.0 & 53.5 & 30.9 & 19.7 & 10.9 & 11.5 & ~6.2 \\
Emuru & $\text{CER}_{0.15}$ & ~7.4K & 67.7 & 34.3 & 28.9 & 24.6 & 19.6 & 16.1 & ~7.9 & ~6.3 & ~5.2 \\
Emuru & $\text{CER}_{0.30}$ & 23.3K & 62.4 & \underline{\textbf{21.7}} & \underline{\textbf{22.9}} & \underline{\textbf{16.5}} & \underline{\textbf{14.3}} & 10.6 & \underline{~7.7} & ~5.9 & \underline{~4.0} \\
Emuru & $\text{CER}_{0.45}$ & 39.8K & 65.5 & 24.9 & 23.9 & 21.3 & 14.7 & 11.5 & ~7.8 & ~6.0 & ~5.5 \\
Emuru & $\text{CER}_{0.60}$ & 53.3K & 66.2 & 29.6 & 29.9 & 18.8 & 15.2 & 11.5 & ~7.9 & ~6.2 & ~5.1 \\
Emuru & -                   & 87.8K & 66.7 & 51.7 & 58.0 & 51.4 & 35.6 & 34.0 & 11.8 & 12.1 & ~6.8 \\
\midrule
DiffPen & $\text{HWD}_{25\%}$ & 22.0K & 80.5 & 54.4 & 42.2 & 30.1 & 23.5 & 15.8 & ~10.0 & \underline{\textbf{~4.5}} & \underline{\textbf{~3.9}} \\
DiffPen & $\text{HWD}_{50\%}$ & 43.9K & 76.0 & 41.9 & 31.5 & 22.1 & 17.3 & 10.7 & ~7.5 & ~5.2 & ~4.5 \\
DiffPen & $\text{HWD}_{75\%}$ & 65.9K & 83.4 & 56.9 & 49.8 & 33.8 & 24.8 & 14.9 & ~9.8 & ~8.3 & ~6.3 \\
DiffPen & $\text{CER}_{0.15}$ & ~0.2K & ~~~- & ~~~- & ~~~- & ~~~- & 98.9 & 12.4 & ~7.4 & ~5.6 & ~4.5 \\
DiffPen & $\text{CER}_{0.30}$ & ~3.5K & ~~~- & 44.4 & 36.7 & 27.4 & 24.7 & 15.7 & ~8.4 & ~5.4 & ~4.5 \\
DiffPen & $\text{CER}_{0.45}$ & 19.7K & \underline{74.8} & 46.7 & 42.0 & 25.8 & 23.2 & 14.4 & ~9.0 & ~4.7 & ~4.0 \\
DiffPen & $\text{CER}_{0.60}$ & 51.8K & 80.2 & 53.3 & 43.0 & 26.7 & 22.9 & 14.4 & ~9.0 & ~7.3 & ~5.6 \\
DiffPen & -                   & 87.8K & ~~~- & \underline{36.7} & \underline{28.4} & \underline{21.8} & \underline{15.7} & \underline{10.1} & \underline{~7.2} & ~4.7 & ~4.6 \\
\midrule
VATr++ & $\text{HWD}_{25\%}$ & 22.0K & 96.9 & 59.4 & 49.0 & \underline{30.7} & \underline{24.1} & \underline{15.9} & 10.2 & ~5.1 & ~4.6 \\
VATr++ & $\text{HWD}_{50\%}$ & 43.9K & 94.7 & 63.6 & 59.6 & 33.8 & 24.9 & 16.1 & ~9.5 & ~9.6 & ~4.3 \\
VATr++ & $\text{HWD}_{75\%}$ & 65.9K & \underline{89.2} & 65.0 & 69.0 & 38.3 & 27.3 & 18.4 & 11.9 & 11.2 & ~4.2 \\
VATr++ & $\text{CER}_{0.15}$ & ~9.9K & 94.6 & \underline{49.2} & \underline{40.1} & 31.2 & 25.8 & 16.9 & \underline{\textbf{~6.8}} & \underline{~5.0} & ~4.4 \\
VATr++ & $\text{CER}_{0.30}$ & 43.8K & 94.3 & 64.5 & 57.7 & 40.4 & 26.7 & 17.0 & 11.5 & 10.3 & \underline{~4.1} \\
VATr++ & $\text{CER}_{0.45}$ & 74.2K & 92.0 & 66.3 & 62.9 & 37.8 & 28.0 & 19.0 & 11.6 & 11.2 & ~4.4 \\
VATr++ & $\text{CER}_{0.60}$ & 85.4K & 95.7 & 67.0 & 60.9 & 37.4 & 30.2 & 17.6 & 11.2 & 11.6 & ~6.5 \\
VATr++ & -                   & 87.8K & 95.9 & 66.4 & 61.0 & 37.9 & 29.3 & 18.5 & 11.3 & 11.5 & ~6.6 \\
\bottomrule
\end{tabular}
}
\caption{CER scores obtained by pretraining DefCRNN on the generated data and then fine-tuned on different portions of the Leopardi dataset. Bold indicates the best overall score for each fine-tuning setting; underline indicates the best score within each setting (HTG model and filtering strategy).}
\label{tab:cer_leopardi}\vspace{-0.5cm}
\end{table}

\begin{table}[]
\centering
\setlength{\tabcolsep}{.3em}
\resizebox{\linewidth}{!}{
\begin{tabular}{llc ccccccccc}
\toprule
    & & & \multicolumn{9}{c}{\textbf{CER ($\times$100)}} \\
\cmidrule{4-12}
\textbf{Data} & \textbf{Filter} & \textbf{\#samples} & \textbf{0\%} & \textbf{1.25\%} & \textbf{2.5\%} & \textbf{5\%} & \textbf{10\%} & \textbf{25\%} & \textbf{50\%} & \textbf{75\%} & \textbf{100\%} \\
\midrule
Real & - & ~0.5K & ~~~- & ~~~- & ~~~- & ~~~- & ~~~- & ~~~- & ~6.5 & ~5.1 & ~3.6 \\
\midrule
Emuru & $\text{HWD}_{25\%}$ & ~5.7K & 18.7 & 17.6 & 15.3 & 13.7 & 10.8 & ~7.5 & ~5.7 & ~5.5 & ~5.1 \\
Emuru & $\text{HWD}_{50\%}$ & 11.4K & 18.8 & 16.7 & 14.4 & 13.5 & 10.1 & ~6.8 & ~6.2 & ~5.5 & ~5.5 \\
Emuru & $\text{HWD}_{75\%}$ & 17.1K & 15.9 & 13.5 & 13.5 & 13.3 & \underline{\textbf{~8.7}} & ~5.8 & ~9.2 & ~8.3 & ~7.0 \\
Emuru & $\text{CER}_{0.15}$ & 16.7K & 15.8 & 13.4 & 12.5 & 12.2 & 10.3 & ~6.2 & ~5.1 & ~5.6 & ~4.3 \\
Emuru & $\text{CER}_{0.30}$ & 20.4K & \underline{\textbf{15.3}} & 15.2 & \underline{\textbf{11.9}} & \underline{\textbf{10.3}} & ~9.5 & ~6.3 & \underline{\textbf{~4.4}} & \underline{\textbf{~4.2}} & \underline{\textbf{~3.5}} \\
Emuru & $\text{CER}_{0.45}$ & 21.5K & 16.6 & 13.9 & 13.4 & 11.1 & ~9.8 & ~6.7 & ~6.1 & ~5.2 & ~4.7 \\
Emuru & $\text{CER}_{0.60}$ & 21.9K & 17.1 & 14.5 & 13.2 & 12.1 & ~9.7 & \underline{\textbf{~5.0}} & ~6.3 & ~5.7 & ~4.1 \\
Emuru & -                   & 23.1K & 16.1 & \underline{\textbf{12.9}} & 12.4 & 11.2 & ~9.9 & ~6.6 & ~5.0 & ~4.8 & ~4.0 \\
\midrule
DiffPen & $\text{HWD}_{25\%}$ & ~5.8K & ~~~- & 32.7 & 24.4 & 18.8 & 51.1 & 12.9 & ~7.1 & ~6.2 & ~4.2 \\
DiffPen & $\text{HWD}_{50\%}$ & 11.6K & 72.9 & 31.8 & 23.3 & 20.5 & 13.9 & ~8.3 & ~7.5 & ~6.2 & \underline{~3.9} \\
DiffPen & $\text{HWD}_{75\%}$ & 17.3K & 72.6 & 29.6 & 24.7 & 19.9 & \underline{11.6} & ~7.2 & ~8.1 & ~6.3 & ~5.1 \\
DiffPen & $\text{CER}_{0.15}$ & 13.8K & 68.7 & \underline{28.1} & \underline{21.7} & 18.3 & 13.4 & ~9.8 & ~6.9 & ~5.0 & ~4.8 \\
DiffPen & $\text{CER}_{0.30}$ & 20.3K & 68.3 & 30.5 & 23.8 & 19.5 & 14.4 & ~9.0 & ~6.7 & ~6.1 & ~5.2 \\
DiffPen & $\text{CER}_{0.45}$ & 22.4K & 78.0 & 29.1 & 23.4 & 19.3 & 13.8 & ~7.2 & ~7.4 & ~6.4 & ~4.7 \\
DiffPen & $\text{CER}_{0.60}$ & 22.9K & 72.2 & 32.0 & 25.4 & 20.7 & 12.4 & \underline{~7.0} & ~8.9 & ~6.8 & ~6.3 \\
DiffPen & -                   & 23.1K & \underline{67.0} & 37.8 & 41.9 & \underline{17.1} & 13.7 & ~8.2 & \underline{~5.1} & \underline{~4.6} & ~4.1 \\
\midrule
VATr++ & $\text{HWD}_{25\%}$ & ~5.8K & ~~~- & \underline{41.9} & 31.3 & 23.2 & 51.8 & ~9.9 & ~8.4 & ~6.2 & \underline{~3.9} \\
VATr++ & $\text{HWD}_{50\%}$ & 11.6K & 86.8 & 46.1 & 29.7 & 22.5 & 16.0 & \underline{~9.0} & ~6.5 & ~5.3 & ~4.1 \\
VATr++ & $\text{HWD}_{75\%}$ & 17.3K & 81.7 & 45.4 & 35.9 & 27.2 & 14.8 & 10.2 & ~8.4 & ~6.3 & ~5.4 \\
VATr++ & $\text{CER}_{0.15}$ & 22.0K & \underline{78.3} & 64.3 & \underline{24.7} & 20.3 & 16.4 & \underline{~9.0} & \underline{~6.2} & ~5.7 & \underline{~3.9} \\
VATr++ & $\text{CER}_{0.30}$ & 23.0K & 83.0 & 62.4 & 28.2 & \underline{17.8} & \underline{13.7} & 10.3 & ~6.3 & \underline{~5.0} & ~5.0 \\
VATr++ & $\text{CER}_{0.45}$ & 23.1K & 81.0 & 45.4 & 34.5 & 27.4 & 14.7 & ~9.1 & ~8.2 & ~6.4 & ~6.3 \\
VATr++ & $\text{CER}_{0.60}$ & 23.1K & 81.9 & 44.5 & 34.5 & 28.9 & 14.7 & 10.4 & ~8.2 & ~6.2 & ~5.4 \\
VATr++ & -                   & 23.1K & 84.8 & 47.6 & 39.2 & 28.8 & 15.9 & ~9.2 & ~8.1 & ~7.2 & ~5.5 \\
\bottomrule
\end{tabular}
}
\caption{CER scores obtained by pretraining DefCRNN on the generated data and then fine-tuned on different portions of the Washington dataset. Bold indicates the best overall score for each fine-tuning setting; underline indicates the best score within each setting (HTG model and filtering strategy).}
\label{tab:cer_washington}\vspace{-0.2cm}
\end{table}

\begin{table}[]
\centering
\setlength{\tabcolsep}{.3em}
\resizebox{\linewidth}{!}{
\begin{tabular}{llc ccccccccc}
\toprule
    & & & \multicolumn{9}{c}{\textbf{CER ($\times$100)}} \\
\cmidrule{4-12}
\textbf{Data} & \textbf{Filter} & \textbf{\#samples} & \textbf{0\%} & \textbf{1.25\%} & \textbf{2.5\%} & \textbf{5\%} & \textbf{10\%} & \textbf{25\%} & \textbf{50\%} & \textbf{75\%} & \textbf{100\%} \\
\midrule
Real & - & ~0.5K & ~~~- & ~~~- & ~~~- & ~~~- & ~~~- & ~~~- & ~~~- & ~7.1 & ~5.0 \\
\midrule
Emuru & $\text{HWD}_{25\%}$ & 17.3K & 37.1 & 12.2 & 10.9 & ~9.4 & \underline{\textbf{~6.7}} & ~6.7 & ~6.9 & ~6.3 & ~6.0 \\
Emuru & $\text{HWD}_{50\%}$ & 34.6K & 29.0 & 13.7 & 11.6 & 10.8 & ~7.5 & ~6.8 & ~7.0 & ~6.9 & ~6.6 \\
Emuru & $\text{HWD}_{75\%}$ & 51.9K & 26.6 & 29.0 & 28.4 & 11.0 & ~7.2 & ~7.5 & ~6.7 & ~5.8 & ~5.5 \\
Emuru & $\text{CER}_{0.15}$ & ~5.7K & 20.0 & 11.9 & 10.0 & ~9.0 & ~7.6 & ~6.9 & ~6.5 & ~6.4 & ~6.2 \\
Emuru & $\text{CER}_{0.30}$ & 22.8K & 21.6 & 12.6 & 11.5 & ~9.6 & ~7.8 & ~6.9 & ~6.3 & ~6.1 & ~5.9 \\
Emuru & $\text{CER}_{0.45}$ & 37.6K & 21.5 & 13.2 & 10.9 & ~9.7 & ~7.5 & ~6.6 & ~6.6 & ~5.9 & ~5.7 \\
Emuru & $\text{CER}_{0.60}$ & 46.3K & \underline{\textbf{18.1}} & 26.2 & 25.4 & 10.1 & ~7.2 & ~7.2 & ~6.7 & ~6.4 & ~6.4 \\
Emuru & -                   & 70.5K & 23.0 & \underline{\textbf{10.3}} & \underline{\textbf{~8.8}} & \underline{\textbf{~8.1}} & ~6.8 & \underline{\textbf{~6.4}} & \underline{~5.7} & \underline{~5.7} & \underline{~5.4} \\
\midrule
DiffPen & $\text{HWD}_{25\%}$ & 17.6K & 59.1 & 27.8 & 21.2 & 16.4 & ~9.6 & \underline{~7.4} & ~7.5 & ~6.0 & ~6.4 \\
DiffPen & $\text{HWD}_{50\%}$ & 35.2K & 62.4 & 27.7 & 21.2 & 15.6 & 10.0 & ~9.0 & ~7.8 & ~6.2 & ~5.8 \\
DiffPen & $\text{HWD}_{75\%}$ & 52.9K & 64.5 & 35.6 & 27.0 & 20.0 & \underline{~8.0} & ~7.7 & ~7.9 & ~6.4 & ~6.5 \\
DiffPen & $\text{CER}_{0.15}$ & ~4.9K & \underline{54.7} & 56.2 & 56.2 & 21.6 & ~8.6 & ~8.5 & 10.5 & ~8.9 & ~5.3 \\
DiffPen & $\text{CER}_{0.30}$ & 25.3K & 59.2 & 31.4 & 24.2 & 17.4 & \underline{~8.0} & ~7.7 & ~7.7 & ~7.0 & ~6.3 \\
DiffPen & $\text{CER}_{0.45}$ & 49.3K & 65.3 & 32.8 & 24.0 & 18.7 & ~9.7 & ~8.4 & ~8.4 & ~7.8 & ~8.0 \\
DiffPen & $\text{CER}_{0.60}$ & 63.5K & 64.2 & 37.9 & 29.5 & 21.1 & ~8.2 & ~9.9 & ~8.2 & ~7.2 & ~6.9 \\
DiffPen & -                   & 70.5K & 69.4 & \underline{18.6} & \underline{15.8} & \underline{12.6} & ~8.9 & \underline{~7.4} & \underline{\textbf{~5.5}} & \underline{\textbf{~5.1}} & \underline{\textbf{~4.8}} \\
\midrule
VATr++ & $\text{HWD}_{25\%}$ & 17.6K & \underline{87.7} & 36.8 & 26.0 & 19.8 & \underline{~9.4} & ~9.1 & \underline{~7.6} & \underline{~7.0} & \underline{~6.0} \\
VATr++ & $\text{HWD}_{50\%}$ & 35.2K & 94.2 & 37.6 & 26.2 & 20.0 & 10.3 & ~9.5 & ~9.3 & ~8.1 & ~9.1 \\
VATr++ & $\text{HWD}_{75\%}$ & 52.9K & 88.7 & 56.6 & 43.1 & 29.6 & ~9.6 & 10.4 & ~9.3 & ~8.2 & ~7.6 \\
VATr++ & $\text{CER}_{0.15}$ & ~1.9K & ~~~- & ~~~- & ~~~- & ~~~- & ~~~- & ~9.9 & ~9.5 & ~7.4 & ~8.0 \\
VATr++ & $\text{CER}_{0.30}$ & 18.0K & 93.4 & \underline{31.8} & \underline{22.5} & \underline{19.2} & ~9.7 & \underline{~8.2} & ~9.0 & ~7.3 & \underline{~6.0} \\
VATr++ & $\text{CER}_{0.45}$ & 47.7K & 92.1 & 64.6 & 60.8 & 45.0 & \underline{~9.4} & ~9.3 & ~8.8 & ~7.5 & ~6.6 \\
VATr++ & $\text{CER}_{0.60}$ & 65.6K & 91.9 & 71.2 & 75.2 & 50.4 & \underline{~9.4} & ~9.2 & ~9.6 & ~9.0 & ~8.3 \\
VATr++ & -                   & 70.5K & 87.9 & 66.9 & 60.1 & 62.5 & 11.1 & ~9.0 & 11.2 & 10.1 & ~10.0 \\
\bottomrule
\end{tabular}
}
\caption{CER scores obtained by pretraining DefCRNN on the generated data and then fine-tuned on different portions of the Saint Gall dataset. Bold indicates the best overall score for each fine-tuning setting; underline indicates the best score within each setting (HTG model and filtering strategy).}
\label{tab:cer_saintgall}
\end{table}

\tit{Effect of Filtering}
Finally, we consider the effect of filtering with the two proposed CER-based and HWD-based strategies (as observed from~\Cref{tab:cer_leopardi,tab:cer_saintgall,tab:cer_washington}). 
Notably, no clear trend emerges between HTR performance and filtering based on handwriting style similarity (HWD). This suggests that strictly enforcing style consistency between the synthetic and real datasets does not necessarily lead to better recognition performance.
Conversely, filtering based on CER appears to have a more direct impact, as observed in~\cite{nikolaidou2024rethinking}. 
The best-performing configurations are typically those where the filtering threshold is set to $\text{CER}_{0.30}$ or when no filtering is applied at all. 
This suggests that the amount of training samples is a more important factor than their quality. 
A weak filter removes the worst examples while keeping enough variety in the data, while a strict filter may remove too many samples and hurt performance. 
Moreover, a too-strict CER-based filtering could remove too many samples, preventing the HTR model from converging (as in the case of pretraining on DiffPen-generated data for Leopardi with a filter with $\text{CER}_{0.15}$). 
For these reasons, not filtering can yield better results, as it maximizes variability in the pretraining data, improving the HTR model's generalizability.  

\section{Conclusions}
\label{sec:conclusion}
We have explored low-resource HTR on historical manuscripts and proposed a pipeline leveraging synthetic data generated by state-of-the-art HTG models to pre-train an HTR model, which is then fine-tuned on a few real samples.
Our findings give the following insights on several studied aspects, such as \textit{style fidelity}, \textit{style variability}, and \textit{synthetic data filtering}.

\tit{Style Fidelity} 
Among the evaluated HTG models, Emuru consistently generates the most style-faithful handwriting samples, leading to superior zero-shot HTR performance, indicating the importance of style fidelity in zero-shot scenarios and the importance of incorporating variability in the training of HTG models in order to generalize to new styles.

\tit{Style Variability} 
When only a small number of real images (fewer than 130) are available for fine-tuning, pretraining on data that closely matches the target handwriting style leads to better recognition results. 
However, if more real samples are available, diversity in the pretraining set becomes more important than style similarity. In these cases, training on the more varied DiffPen-generated datasets leads to better generalization and improved HTR performance.
This shows the effect of style diversity depending on the amount of fine-tuning data and the importance of data variability in order to have efficient HTR systems.

\tit{Synthetic Data Filtering} 
We have observed no clear benefit from filtering based on HWD. 
In contrast, filtering based on CER with a too strict threshold can lead to removing too many useful training samples or even hinder the HTR model convergence.
However, removing noisy data with false content in a more relaxed scenario might contribute to the HTR performance.
In designing filtering strategies, a balanced threshold seems most beneficial, aiming to discard only highly erroneous samples while preserving style variability that is crucial to the pretraining process.\newline

To conclude, this study aimed to offer insights into the current state of HTG research and its potential contributions to enhancing low-resources HTR in low-resources scenarios.
These insights contribute to a deeper understanding of how synthetic data from HTG models can best be leveraged for HTR, providing practical guidelines for selecting and refining pretraining datasets in low-resource settings. 
More broadly, this study sheds light on the maturity of current HTG techniques and their potential to enhance HTR in challenging historical manuscript collections.
By shedding light on the existing HTG capabilities, we aim to help the design of novel HTG models for boosting HTR in low-resource scenarios.

\section*{Acknowledgements}
This work was supported by the ``AI for Digital Humanities'' project (Pratica Sime n.2018.0390), funded by ``Fondazione di Modena'' and the PNRR project Italian Strengthening of ESFRI RI Resilience (ITSERR) funded by the European Union – NextGenerationEU (CUP: B53C22001770006).

{
    \small
    \bibliographystyle{ieeenat_fullname}
    \bibliography{main}
}

\end{document}